\documentclass[letterpaper]{article} 
\usepackage[submission]{aaai23}  
\usepackage{times}  
\usepackage{helvet}  
\usepackage{courier}  
\usepackage[hyphens]{url}  
\usepackage{graphicx} 
\usepackage{natbib}  
\usepackage{caption} 
\usepackage{algorithm}
\usepackage{algorithmic}

\usepackage{amsmath}
\usepackage{subfigure}
\usepackage{amssymb}
\usepackage{booktabs}
\usepackage{multirow}
\usepackage{graphicx}
\usepackage{color}
\usepackage{enumitem}
\usepackage{listings}
\usepackage{pythonhighlight}
\usepackage{cuted}
\usepackage{newfloat}
\usepackage{listings}

\DeclareCaptionStyle{ruled}{labelfont=normalfont,labelsep=colon,strut=off} 
\floatstyle{ruled}
\newfloat{listing}{tb}{lst}{}
\floatname{listing}{Listing}
\definecolor{deepblue}{rgb}{0,0,0.5}
\definecolor{deepred}{rgb}{0.6,0,0}
\definecolor{deepgreen}{rgb}{0,0.5,0}

\usepackage{listings}

\newcommand\pythonstyle{\lstset{
language=Python,
basicstyle=\ttm,
morekeywords={self},              
keywordstyle=\ttb\color{deepblue},
emph={MyClass,__init__},          
emphstyle=\ttb\color{deepred},    
stringstyle=\color{deepgreen},
frame=tb,                         
showstringspaces=false
}}

\newcommand\pythoninline[1]{{\pythonstyle\lstinline!#1!}}

\title{NoMorelization: Building Normalizer-Free Models from a Sample's Perspective}
\author {
    Chang Liu, Yuwen Yang, Yue Ding, Hongtao Lu$^*$
}
\affiliations {
    Department of computer science and engineering, Shanghai Jiao Tong University\\ 
    \{isonomialiu, youngfish, dingyue, htlu\}@sjtu.edu.cn
}

\begin{document}
\maketitle

\begin{abstract}
The normalizing layer has become one of the basic configurations of deep learning models, but it still suffers from computational inefficiency, interpretability difficulties, and low generality.
After gaining a deeper understanding of the recent normalization and normalizer-free research works from a sample's perspective, we reveal the fact that the problem lies in the sampling noise and the inappropriate prior assumption. 
In this paper, we propose a simple and effective alternative to normalization, which is called ``NoMorelization''. 
NoMorelization is composed of two trainable scalars and a zero-centered noise injector.
Experimental results demonstrate that NoMorelization is a general component for deep learning and is suitable for different model paradigms (e.g., convolution-based and attention-based models) to tackle different tasks (e.g., discriminative and generative tasks).
Compared with existing mainstream normalizers (e.g., BN, LN, and IN) and state-of-the-art normalizer-free methods, NoMorelization shows the best speed-accuracy trade-off.
\end{abstract}

\section{Introduction}
The marriage of skip connection \cite{resnet} and normalization methods \cite{BN,LN,IN,GN} has become the mainstream paradigm and witnessed significantly advanced performance across different domains. The pervasiveness of normalization layers stabilizes the very deep neural networks during training to make modern deep networks training possible, such as Convolution Neural Networks (CNNs) \cite{resnet, convnext} and Transformers \cite{transformer, bert, swin}.

However, normalization methods have three significant practical disadvantages. 
Firstly, normalization is a surprisingly expensive computational primitive with memory overhead \cite{overhead}.
Secondly, despite the pragmatic successes of normalization methods in many fields,  it is still difficult for researchers to interpret the underlying mechanism. Moreover, several works \cite{four, InstanceEB, batch} claimed that by understanding some of the mechanisms, normalization methods could be polished in practice (See Sec. \ref{sec:understand}).
Thirdly, and most importantly, a certain normalization method is usually designed for limited backbones and tasks, while the general normalization layer is still to be designed. 

For example, batch normalization (BN) is common in CNNs for discriminative tasks like classification while inferior to instance normalization (IN) in generative tasks \cite{cycle}, meanwhile layer normalization (LN) is believed to outperform BN in Transformers and recently developed CNNs \cite{convnext}.

There is a surging interest in building normalizer-free methods, and several normalizer-free methods \cite{fixup, skipinit, re0} have shown competitive performance compared with state-of-the-art models, faster speed, and good interpretability. 

Nevertheless, such methods still have minor gaps in accuracy. Some of the underlying complex regularization effects of normalization, especially batch normalization \cite{HowDB}, might be the key to pushing the normalizer-free methods forward a step. NFNets \cite{sota1,sota2} introduced external regularization and developed a specially tailored CNN to outperform state-of-the-art models with BN, firstly. However, external regularization is costly and slows down the training process.

\begin{figure}[htbp]
\centering 
    \centering
    \includegraphics[width=0.9\linewidth]{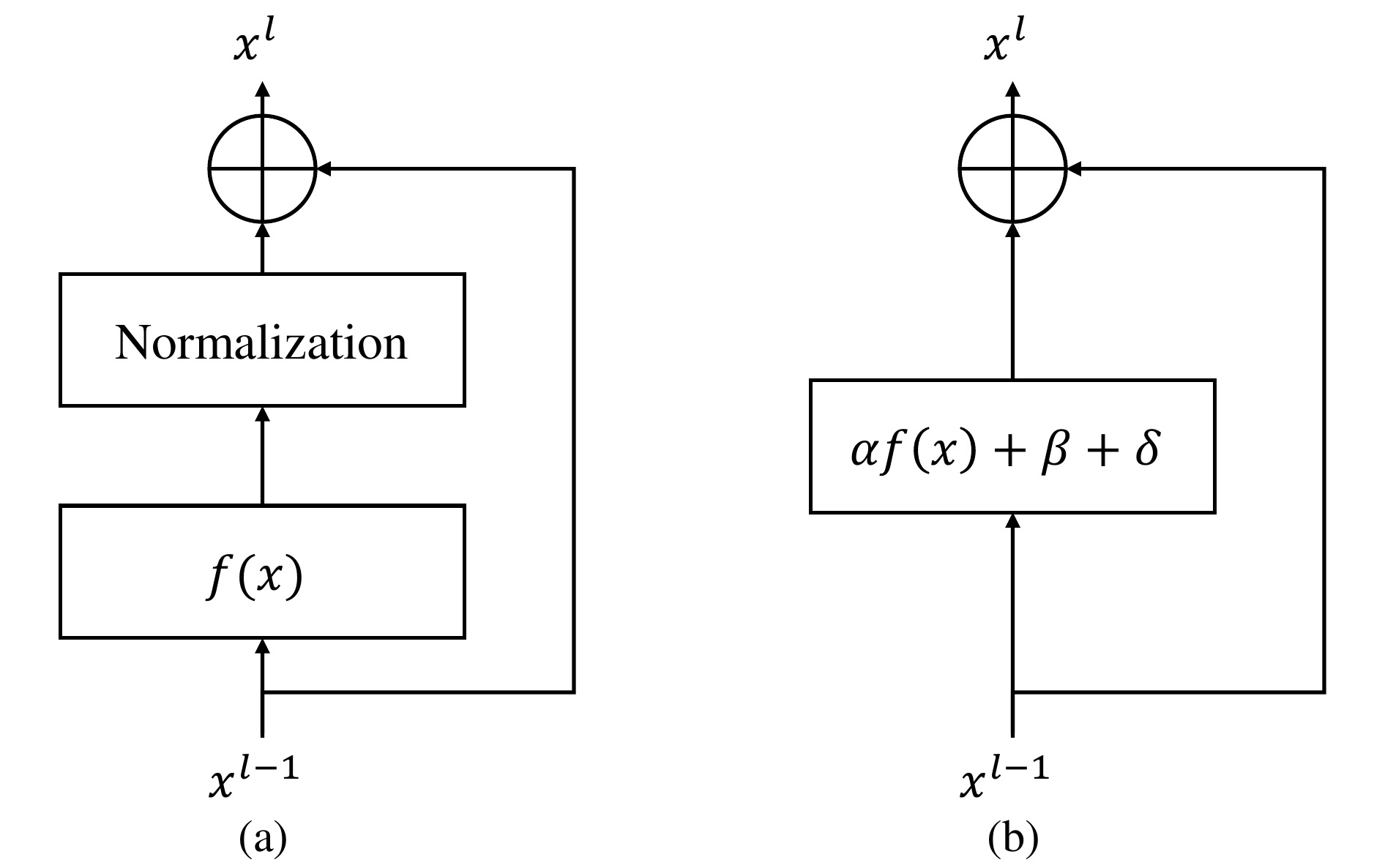}
    \caption{(a) A residual block with normalization layers, where $f(\cdot)$ denotes core computational primitive (\textit{e.g.}, convolution, attention, and aggregation).
    (b) NoMorelization replaces normalization layers with a learnable scalar-multiplier $\alpha$, bias $\beta$ and a zero-centered noise injector $\delta$.} 
    \label{fig:NoMore} 
\end{figure}

 In this work, we aim to confront the challenge of building an \textit{efficient}, \textit{explainable}, and \textit{general} normalizer-free module, dubbed as \textbf{NoMorelization} shown in Fig. \ref{fig:NoMore}. 

 NoMorelization shows a better effectiveness-efficiency trade-off than existing mainstream normalization layers and normalizer-free methods.
We build NoMorelization by explaining BN from a sample's perspective, \textit{i.e.,} down-scaling the residual path and an \textbf{additional noise regularization}. What's more, NoMorelization is a general component for different backbones (CNN and Transformer) and tasks (discriminative and generative) to substitute different normalizers (BN, LN, and IN). 

Our main contributions are as follows:
\begin{itemize}[leftmargin=*]
	\item We find the regularization that current normalizer-free methods lack, namely noise injection from a sample's perspective. 
	By correcting the erroneous assumptions about the distribution of features, we model and derive the nature of injected noise and experimentally verify our assumptions. 
	\item We propose a general normalizer-free method called NoMorelization. NoMorelization is composed of learnable scalars on the residual branch and a zero-centered Gaussian noise injector during training.
	\item  NoMorelization outperforms mainstream normalizers and state-of-the-art normalizer-free methods in multiple backbones and tasks with better speed-accuracy trade-offs.
\end{itemize}

\section{Related Works}
	\subsection{Understanding Normalization Layers}
	\label{sec:understand}
	\subsubsection{Normalization Layers}
	Normalization layers standardize given input tensor $\boldsymbol{x}_i$ by Eq.\ref{eq:Norm}, i.e., minus input's mean and divide it by its standard deviation (with a small positive number $\epsilon$). Then an optional affine transformation with a learnable mean $\beta$ and standard deviation $\gamma$.
	\begin{equation}
		\label{eq:Norm}
		\hat{\boldsymbol{x}}_{i}=\gamma \frac{\boldsymbol{x}_{i}-\mu}{\sqrt{\sigma^{2}+\epsilon}}+\beta
	\end{equation}
	
	Several popular variants of normalization layers have been prevailing since their origin, including Batch Normalization (BN) \cite{BN}, Layer Normalization (LN)  \cite{LN}, Instance Normalization (IN)  \cite{IN}, and Group Normalization (GN) \cite{GN}. The main difference among them is reflected in the statistics $\mu$ and $\sigma^2$.

	\subsubsection{Properties of Normalization Layers}
	Depending on the calculated statistics, Normalization layers have different properties. 
	Different properties will cause different advantages and disadvantages,
	 which is why it is difficult to design a general normalization layer.
	\begin{itemize}[leftmargin=*]
		\item \textbf{Global/Adaptive statistics.} According to the way to gather, statistics can be divided into global and adaptive statistics. Global statistics are consistent for different samples and can be obtained through the moving average during training. In contrast, the statistics are unstable due to batch dependence \cite{ProxyNorm}. The adaptive statistics are related to the sample, so the adaptive statistics must be gathered during inference. BN uses global statistics, while other methods (LN, IN, etc.) use adaptive statistics. Global statistics of BN can cause a training-inference discrepancy and lead to a decrease in model performance \cite{four, batch}.
		\item \textbf{Unexpectedly Luxury} Normalization is a surprisingly expensive computational primitive. It can occupy 25\% of the total training time of a ResNet model \cite{slow} and cause memory overhead \cite{overhead} for computing statistics.
		\item \textbf{Complex Regularization} It is widely believed that batch normalization also acts as a regularizer with noise injection \cite{TowardsUR,InstanceEB}. We can finetune the intensity of regularization by changing batch size.  
		The smaller the batch size, the greater the intensity of regularization. In addition, researchers have found that normalization can smooth the lost landscape \cite{understanding},
	 making the model have a certain degree of scaling invariance \cite{IN}, bias the model to its shallow path \cite{shallow}, orthogonalize the representations \cite{Orthogonalize},  increases adversarial vulnerability \cite{Vulnerability}, and so on.

	\end{itemize}

	\subsection{Investigating Normalizer-Free (NF) Methods}
	There has been surging interest in designing normalizer-free methods. Through delicate initialization \cite{fixup}, learnable scalar \cite{skipinit, re0}, and scaled weight standardization \cite{sota1},  normalizer-free ResNets shows competitive results with BN ResNets. Recently \cite{sota2} proposed a new NF backbone, namely NFNet. With adaptive gradient clipping, NFNet can outperform SOTA CNN-based models in ImageNet classification.
	\subsubsection{The Power of Down-scaling}
	 Normalization is indispensable in modern residual blocks due to its ability to down-scale. Let $\boldsymbol{x}^{l}$ denotes the input batch of the $l$-th residual block, and $\boldsymbol{x}^1$  denotes the input of the model. A common assumption is that each sample of the network's input is independently and identically distributed with Gaussian mean zero variance 1.
	 \begin{equation}
	     \mathbf{E}\left(\boldsymbol{x}_i^{1}\right)=0, \mathbf{Var}\left(\boldsymbol{x}_{i}^{1}\right)=1,
	 \end{equation}
	 where the ${x}_{i}^{1}$ denotes the $i$-th sample of the input batch. The forward pass of $l$-th residual block is
	 \begin{equation}
	     \boldsymbol{x}^{l+1} = f^l(\boldsymbol{x}^l)+\boldsymbol{x}^l,
	 \end{equation}
	 where $f^l(\cdot)$ is the composition of the layers and activation functions within the $l$-th block. With widely used ReLU activation \cite{relu} and initialization method (such as Kaiming Init \cite{Heinit}), the layer outputs $f^{l}\left(x^{l}\right)$ are independent of their inputs, and thus the activations’ variance of an Unnormalized network grows exponentially with the number of blocks.
	 \begin{equation}
	     \mathbf{Var}\left(x_{i}^{l+1}\right)=\mathbf{Var}\left(x_{i}^{l}\right)+\mathbf{Var}\left(f_{i}^{l}\left(x^{l}\right)\right) \approx	 2^l
	 \end{equation}
	  Exponentially increasing activation variance can cause exploding gradients at the very beginning of training. But with normalization, the activation variance will be down-scaled to grow linearly.
	  	\begin{equation}
	     \mathbf{Var}\left(x_{i}^{l+1}\right)=\mathbf{Var}\left(x_{i}^{l}\right)+\mathbf{Var}\left(\operatorname{Norm}(f_{i}^{l}\left(x^{l})\right)\right) = l+1
	 \end{equation}
    Therefore, existing NF works focus on stabilizing the outputs' variance by initializing or introducing multipliers.
	 \begin{equation}
	     \boldsymbol{x}^{l+1} = \alpha \times f^l(\boldsymbol{x}^l)+\boldsymbol{x}^l,
	     \label{eq:skipinit}
	 \end{equation}
	 where the $\alpha$ is a trainable parameter and initialized as 0. This multiplier helps the model to be normalized as an identity network and makes NF networks training possible.
	\subsubsection{Additional Regularization Effects}
	 Despite the achievement in down-scaling, the NF models still suffer some accuracy loss than their Siamese with normalization. This is believed to be due to the additional regularization effect of normalization layers. NFNet \cite{sota2} is the first NF model that achieves SOTA against the normalized. It claims that batch normalization can keep the model outputs' mean to 0 and stabilize the numerical range of the gradient. To this end, NFNet implements scaled weight normalization and adaptive gradient clipping to the NF model.

\section{NoMorelization: A Sample’s Perspective} 
	The key motivation of NoMorelization is to rethink BN from a sample's perspective and thus reveal the regularization effect REALLY MISSED in existing NF methods. The state-of-the-art NFNets \cite{sota2} treat BN's regularization as multiple complex regularization effects. However, from a sample's perspective, we conclude that a very simple noise injector can be implemented as a surrogacy for BN to realize better normalizer-free modules. In the following sections, we empirically illustrate the irrationality of the regularization scheme of NFNet and provide a theoretical basis for the correctness of Noise Injection.
	\subsection{Why Noise Injection}
	\label{sec:why}
	\subsubsection{Existing choices are not reasonable}
\begin{table}
\centering
\small
\resizebox{.95\columnwidth}{!}{
\begin{tabular}{cccc}
\toprule
     & \begin{tabular}[c]{@{}c@{}}W/o \\ Regularization\end{tabular} & \begin{tabular}[c]{@{}c@{}}NF-\\ Regularization\end{tabular} & Improvement   \\ \midrule
BN-Net & 92.47 & 93.23 & 0.76 \\ 
NF-Net & 92.00 & 92.51 & 0.51\\ 
$\Delta$Acc   & 0.47 & 0.72 & \textbf{0.25 ($\ge$0)}          \\ \bottomrule
\end{tabular}}
    \caption{The average accuracy improvement of the regularization in NFNet \cite{sota2} is compared in the CIFAR-10 dataset of ResNet-110 with and without BN. The results show that the BN-Net receives a more noticeable improvement via NFNet's regularization.}
		\label{table: regular}
\end{table}
	We implement two 110-layer ResNets. The first ResNet is implemented with BN and called BN-Net. The multiplier replaces the BN layers of the other ResNet described as Eq. (\ref{eq:skipinit}). 
	This network is called NF-Net. We then perform two sets of training on these two networks. The first set of training does not use any regularization for both networks. The second set is trained with the regularization used by the NFNets \cite{sota2} (\textit{i.e.,} adaptive gradient clipping and scaled weight standardization). The average results are shown in Tab. \ref{table: regular}. It is worth noting that NF-Net with additional NFNets-style regularization can surpass the performance of vanilla BN-Net, but the regularization can improve BN-Net even more. This intuitively shows that \textbf{the existing regularization choice is a complement instead of a substitute for the BN regularization effect}. Similar results can be found in more models and datasets in Tab. \ref{table:class}.
	\begin{figure}
		\centering
		\includegraphics[width=0.8\linewidth]{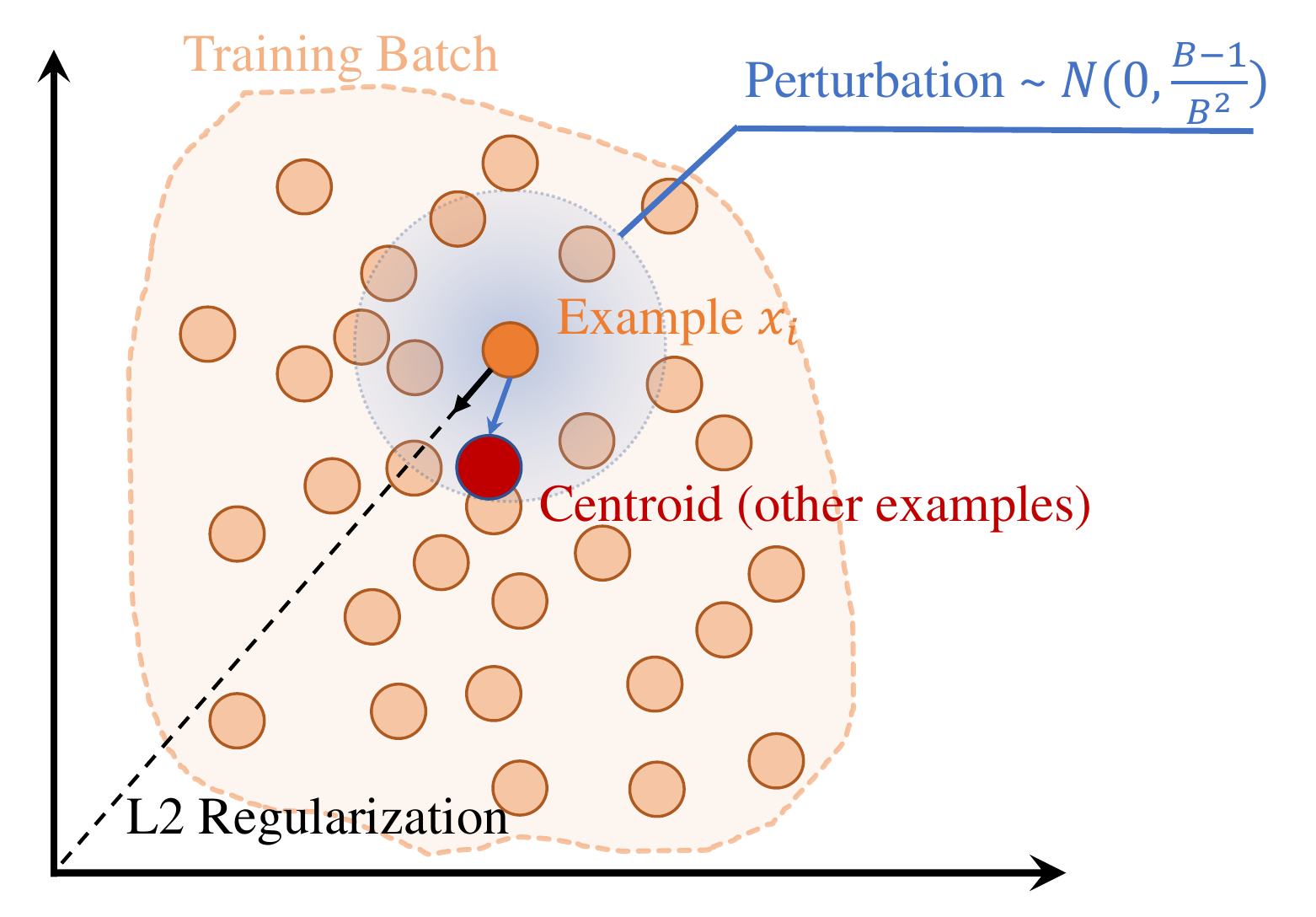}
		\caption{Batch normalization: a sample's perspective (under the I.I.D. assumption)}
		\label{fig:BNAEP}
	\end{figure}
	\subsubsection{Injected Noise Exists} 
	
		Ignoring the learnable parameters(i.e. $\beta$ and $\gamma$) in Eq. \ref{eq:Norm}, we \textbf{deduce what BN has actually done from the perspective of the i-th sample $\boldsymbol{x}_i$}.
	The previous work \cite{skipinit} has proved that the normalization methods can stabilize the variance of the output data. We assume that the standard deviation of the denominator can be discarded together with the learnable standard deviation $\gamma$ for brevity. To simplify the discussion, we consider the case where variables such as $x_i$ are scalars. 
	\begin{equation}
		\label{eq:noise}
		\begin{aligned}
			\hat{x_{i}} &=x_{i}-\mu \\
			&= x_{i}-\frac{1}{B} \sum_{j=1}^{B} x_{j}\\
			&= x_{i}-\frac{1}{B} x_{i}-\frac{1}{B} \sum_{j=1,j \neq i}^{B} x_{j} \\
			&=\frac{B-1}{B} x_{i}-\frac{1}{B}  \sum_{j=1,j \neq i}^{B} x_{j},\\
		\end{aligned}
	\end{equation}
	where $x$ are random variables that obey the standard normal distribution (via the definition of BN). Wherefore $\sum_{j=0}^{B-1} x_{j} \sim \mathcal{N}(0, B-1)$. 
	\begin{equation}
		\label{eq:naive}
		\hat{x_i} = \frac{B-1}{B}x_{i} - \delta, (\delta\sim \mathcal{N}(0, \frac{B-1}{B^2})).
	\end{equation}

	The above formula is the so-called \textbf{a sample's perspective in BN}. As shown in Fig. \ref{fig:BNAEP}, BN firstly performs an L2 regularization on this sample (i.e., $\frac{B-1}{B}\boldsymbol{x}_i$), and then injects a random Gaussian perturbation $\delta$.
	The magnitudes of these two regularizing effects are both related to the batch size $B$. The larger the batch size, the smaller the magnitude of the regularization. This also explains the subtle relationship between the batch size and the corresponding test accuracy found in previous studies \cite{Control,TowardsUR}.

\section{Noise Modeling } 
	Although we claim that noise injection is an additional regularization effect of BN, the effects of sampling noise, which are mostly negative, have also been extensively studied. For example, using a fixed virtual batch can eliminate noise and improve performance (especially on generative models) \cite{vbn}. Fixing the proportion of classes in each batch can also reduce noise and improve the performance in various tasks \cite{batch}, e.t.c.
	
		We are concerned about what the noise is and whether we can use it as a regularization.
	With the well-known two contradictory prior knowledge of previous works:
	\begin{itemize}[leftmargin=*]
		\item 	Without BN's extra regularization effects (including noise), the normalizer-free models suffer from a lower accuracy.
		\item 	Because BN has training noise, alleviating the ``training noise'' of BN will improve models' performance \cite{RepresentativeBN}.
	\end{itemize}
	In general, we believe that BN's noise is beneficial to the training of deep learning models, but the noise is too large. The excessively large part of the noise is a polynomial distribution noise related to the sampling result. Removing this noise and leaving only the Gaussian noise with a mean value of zero is more helpful for model training.
	
	Moreover, the existence of excessive noise is an important reason why BN cannot be generalized to all models and tasks. We can build a general NoMorelization by ``distilling'' the excessive noise, i.e., only injecting a small zero-centered noise during training.
	
	\subsection{Rewrite the Prior Distribution}
	We argue that the prior distribution of the input data should not obey an \textit{i.i.d.} Gaussian distribution.
	If the model needs to extract meaningful representations for downstream tasks, the distribution of the extracted representations tends to be polycentric \cite{rich}. BN is also confirmed to require additional corrections when the non-iid situation is more severe \cite{FedBN}, and BN may fail with large variation \cite{bigfail}.
	The data should be composed of multiple Gaussian distributions, and the number of distributions in classification tasks is at least the number of classes. For dataset $\mathcal{D}=\{(x_i,y_i)\}$, $x \sim \mathcal{N}(\mu^{(y)},\sigma^{(y)2})$, where $y_i$ denotes the distribution (e.g. class) index of $i$-th sample.
	
	\subsection{Rewrite the Noise Model}
	With the modified prior distribution, we can rewrite the noise model in Eq. (\ref{eq:noise}) as:
	\begin{equation}
				\label{eq: new noise}
		\begin{aligned}
			\hat{x_{i}} &=\frac{B-1}{B} \tilde{x}_{i}^{y_i} -\delta, \\
			\text{where } \tilde{x}_{i}&\sim \mathcal{N}(\mu^{y_i},  \sigma^{(y_i)2}), \\
			\delta &\sim \mathcal{N}( \underbrace{ \frac{1}{B} \sum_{j=1, j \neq i}^{B} \mu_j^{y_j}}_{\text{Inter-distribution}},\underbrace{\frac{1}{B^2}\sum_{j=1, j \neq i}^{B}\sigma^{(y_j)2}}_{\text{Intra-distribution}}).\\
		\end{aligned}
	\end{equation}
 	The sum of multiple independent Gaussian variables still follows a Gaussian distribution, but we can refer to the mean and variance parts of this Gaussian distribution as the \textbf{Inter-distribution noise} and the \textbf{Intra-distribution noise}, respectively. Now we are going to discuss the nature of rewrited noise in Eq. (\ref{eq: new noise}).
 	\paragraph{Inter-distribution Noise}
 	Observe the mean value of the disturbance $\delta$, which is actually the summation of the expectation of each distribution sample. Because each sample in the batch is independent, we have
 	\begin{equation}
 		\begin{aligned}
 			P\left\{|y=1|=m_{1}, |y=2|=m_{2}, \cdots, |y=n|=m_{n}\right\}=\\
 			\frac{B!}{m_{1} ! m_{2} ! \cdots m_{n} !} p_{1}^{m_{1}} p_{2}^{m_{2}} \ldots p_{n}^{m_{n}},
 		\end{aligned}
 	\end{equation}
 	where $n$ is the number of distributions, and $B$ is the batch size.
 	$|y=i|$ refers to the number of samples belonging to $i-$th distribution, also denoted as $m_i$, and $p_i$ is the probability of the $i$-th distribution being sampled (the same as the ratio of every distribution when the datasets is large enough).
 Each case of the above sampled polynomial distribution corresponds to a mean value of $\sum_{j=1, j \neq i}^{B} \mu_j^{y_j}$.

	Because the mean value of this noise is also a random variable, when sampling time is large enough, it will be more like a Gaussian distribution approaching the same center, just like the previous noise model in Eq. (\ref{eq:naive}). But for each BN sampling for training, the noise is indeed not zero-centered.
	\paragraph{Intra-distribution Noise} After stripping the inter-distribution noise, the remainder becomes a zero-centered Gaussian noise $\mathcal{N}(  0,\frac{1}{B^2}\sum_{j=1, j \neq i}^{B}\sigma^{(y_j)2})$. We call it intra-distribution noise, and we assume that the standard deviation of this distribution is smaller than the range of inter-distribution noise. We will elaborate and verify our claims in subsequent assertions and experiments.
	\paragraph{Special cases}
	The above formula is complicated, but we can propose special cases. For the $i-$th sample in a batch, if all samples except  $i-$th belong to same distribution (no matter what $i$-th sample's distribution is.) The noise term can be simplified as for: $\delta^{(y)} \sim \mathcal{N}( \frac{B-1}{B} \mu^{y},\frac{B-1}{B^2}\sigma^{(y)2})$. It can be vividly shown in Fig. \ref{fig:NoiseModeling}. Samples perturbed by noise exhibit polycentric distributions related to the sampled class.
	\subsection{Assertions}

	\paragraph{Assertion 1: Intra-distribution noise extraction}

	If we fix a sample $x_i$ and sample other data from an arbitrary distribution of $y_i$ to form a batch and then perform BN, we can get the embeddings of $x_i$ represented by
	$\hat{x_{i}}(1),\cdots,\hat{x_{i}}(K)$ after repeated random sampling and applying BN for $K$ times. 
	Then subtract two different $\hat{x_{i}} $ (\emph{e.g.,} $\hat{x_{i}}(3)-\hat{x_{i}}(4)$), the result should be:
	\begin{equation}
		\delta^\prime \sim \mathcal{N}( 0,\frac{2B-2}{B^2}\sigma^{(y)2}).
	\end{equation}

	In fact, $\delta^\prime$ obtained above should also be a zero-centered Gaussian distribution as long as the proportion of the batched samples belonging to each distribution is constant.
	
	\paragraph{Assertion 2: Inter-distribution noise extraction}
	Based on Assertion 1, if the sampled batch data and the fixed sample $x_i$ are from the same distribution $y_i$ after $K$ sampling and BN, the exception of $\hat{x_{i}} $ should be described as:
	\begin{equation}
		E(\hat{x_{i}} )_{\text{self}}= \frac{B-1}{B} \mu^{y_i}-\frac{B-1}{B} \mu^{y_i}=0.
	\label{a2:eq1}
	\end{equation}

	And if not so, the exception is non-zero:
	\begin{equation}
		E(\hat{x_{i}} )_{y}=\frac{B-1}{B} \mu^{y_i}-\frac{B-1}{B} \mu^{y} \neq 0.
	\label{a2:eq2}
	\end{equation}
	
	\paragraph{Assertion 3: The noise decomposition assumption}
	The influence of BN is mainly dominated by inter-distribution noise. When we really calculate the noise caused by BN, 
we can distinguish which distribution the fixed sample is combined with according to the noise.

We validate our assertions by hypothesis test and visualization experiment in Sec. \ref{sec:assertion}. We demonstrate that our assumptions about the prior distribution are more realistic by the three assertions.

	\subsection{NoMorelization}
	After clarifying the noise effect of BN, we give a simple formula for NoMorelization:
	\begin{equation}
	    \alpha f(\boldsymbol{x})+\beta+ \boldsymbol{\delta},
	    \label{eq.nomore}
	\end{equation}
    where the $\alpha$ and $\beta$ are multiplier and offset initialized as 0 like affine transformation in normalization layers in Eq. (\ref{eq:Norm}). $\boldsymbol{\delta}$ is a standard zero-centered Gaussian noise vector.
    In practice, we find that those backbones using BN tend to prefer larger noise, while those using LN and IN prefer smaller noise. So we multiply $\boldsymbol{\delta}$ by a scalar $\gamma$ as a hyperparameter, and we explain the hyperparameter further in Sec. \ref{sec:ablation}.
    \begin{equation}
	    \alpha f(\boldsymbol{x})+\beta+ \gamma \times \boldsymbol{\delta},
	    \label{eq.nomore1}
	\end{equation}
    In summary, by introducing a zero-centered Gaussian noise, our NoMorelization can achieve an elegant trade-off between speed and accuracy, \textit{i.e.,} NoMorelization is a substitute with a low cost and even exceed the performance of various normalization layers in accuracy. 
	Moreover, our experimental results prove that, with nice interpretability, 
	NoMorelization is more substitutable than complementary to the regularization effect of BN compared to existing normalizer-free methods. So we multiply delta by a scalar as a hyperparameter, and we explain the hyperparameter further in the experiments section.
\begin{table*}
\centering
\small
\resizebox{\linewidth}{!}{
\begin{tabular}{ccccccccc}
\toprule
Dataset                        & Backbone   & Norm                                                                    & NF-Norm                                                                    & Fixup                                                                      & SkipInit                                                                   & NFNet                                                                      & Ours                                                                       & NF-Ours                                                                    \\ \hline
\multirow{4}{*}[-1.5ex]{CIFAR-10}       & ResNet     & \begin{tabular}[c]{@{}c@{}}92.31 $\pm$ 0.36\\ (1 $\times$)\end{tabular} & \begin{tabular}[c]{@{}c@{}}93.06 $\pm$ 0.26\\ (0.43 $\times$)\end{tabular} & \begin{tabular}[c]{@{}c@{}}91.32 $\pm$ 0.40\\ (1.03 $\times$)\end{tabular} & \begin{tabular}[c]{@{}c@{}}91.92 $\pm$ 0.32\\ (1.35 $\times$)\end{tabular} & \begin{tabular}[c]{@{}c@{}}92.76 $\pm$ 0.39\\ (0.43 $\times$)\end{tabular} & \begin{tabular}[c]{@{}c@{}}\textit{92.47 $\pm$ 0.35}\\ \textit{(1.22 $\times$)}\end{tabular} & \begin{tabular}[c]{@{}c@{}} \textbf{93.39 $\pm$ 0.30}\\ (0.43 $\times$)\end{tabular} \\
                               & ConvNeXt   & \begin{tabular}[c]{@{}c@{}}88.47 $\pm$ 0.45\\ (1 $\times$)\end{tabular} & \begin{tabular}[c]{@{}c@{}}89.59 $\pm$ 0.33\\ (0.62 $\times$)\end{tabular} & \begin{tabular}[c]{@{}c@{}}\textit{88.75 $\pm$ 0.41}\\ \textit{(1.20 $\times$)}\end{tabular} & \begin{tabular}[c]{@{}c@{}}88.23 $\pm$ 0.44\\ (1.29 $\times$)\end{tabular} & \begin{tabular}[c]{@{}c@{}}89.92 $\pm$ 0.36\\ (0.59 $\times$)\end{tabular} & \begin{tabular}[c]{@{}c@{}}\textit{89.08 $\pm$ 0.50}\\ \textit{(1.13 $\times$)}\end{tabular} & \begin{tabular}[c]{@{}c@{}} \textbf{90.26 $\pm$ 0.40}\\ (0.57 $\times$)\end{tabular} \\
     & Swin & \begin{tabular}[c]{@{}c@{}}94.45 $\pm$ 0.13\\ (1 $\times$)\end{tabular}        & \begin{tabular}[c]{@{}c@{}}94.64 $\pm$ 0.03\\ (0.84 $\times$)\end{tabular}           & \begin{tabular}[c]{@{}c@{}} N/A \\ N/A\end{tabular}           & \begin{tabular}[c]{@{}c@{}}94.23 $\pm$ 0.18\\ (1.11 $\times$)\end{tabular}           & \begin{tabular}[c]{@{}c@{}}94.81 $\pm$ 0.07\\ (0.88 $\times$)\end{tabular}           & \begin{tabular}[c]{@{}c@{}}\textit{94.72 $\pm$ 0.15}\\ \textit{(1.11 $\times$)}\end{tabular}           & \begin{tabular}[c]{@{}c@{}}\textbf{94.85 $\pm$ 0.07}\\ (0.87 $\times$)\end{tabular}  \\ \hline
\multirow{4}{*}[-1.5ex]{CIFAR-100}      & ResNet     & \begin{tabular}[c]{@{}c@{}}72.12 $\pm$ 0.31\\ (1 $\times$)\end{tabular} & \begin{tabular}[c]{@{}c@{}}72.70 $\pm$ 0.36\\ (0.32 $\times$)\end{tabular} & \begin{tabular}[c]{@{}c@{}}\textit{72.41 $\pm$ 0.64}\\ \textit{(1.01 $\times$)}\end{tabular} & \begin{tabular}[c]{@{}c@{}}72.06 $\pm$ 0.60\\ (1.17 $\times$)\end{tabular} & \begin{tabular}[c]{@{}c@{}}72.53 $\pm$ 0.37\\ (0.37 $\times$)\end{tabular} & \begin{tabular}[c]{@{}c@{}}\textit{72.58 $\pm$ 0.41}\\ \textit{(1.05 $\times$)}\end{tabular} & \begin{tabular}[c]{@{}c@{}}\textbf{72.84 $\pm$ 0.36}\\ (0.35 $\times$)\end{tabular} \\
                               & ConvNeXt   & \begin{tabular}[c]{@{}c@{}}66.37 $\pm$ 0.38\\ (1 $\times$)\end{tabular} & \begin{tabular}[c]{@{}c@{}}68.01 $\pm$ 0.48\\ (0.88 $\times$)\end{tabular} & \begin{tabular}[c]{@{}c@{}}66.04 $\pm$ 0.69\\ (1.41 $\times$)\end{tabular} & \begin{tabular}[c]{@{}c@{}}65.11 $\pm$ 1.39\\ (1.17 $\times$)\end{tabular} & \begin{tabular}[c]{@{}c@{}}67.79 $\pm$ 0.23\\ (0.68 $\times$)\end{tabular} & \begin{tabular}[c]{@{}c@{}}\textit{67.20 $\pm$ 0.96}\\ \textit{(1.17 $\times$)}\end{tabular} & \begin{tabular}[c]{@{}c@{}}\textbf{69.05 $\pm$ 0.44}\\ (0.64 $\times$)\end{tabular} \\
     & Swin & \begin{tabular}[c]{@{}c@{}}76.93 $\pm$ 0.17\\ (1 $\times$)\end{tabular}        & \begin{tabular}[c]{@{}c@{}}\textbf{77.34 $\pm$ 0.14}\\ (0.86 $\times$)\end{tabular}           & \begin{tabular}[c]{@{}c@{}} N/A \\ N/A\end{tabular}           & \begin{tabular}[c]{@{}c@{}} 76.79 $\pm$ 0.35\\ (1.11 $\times$)\end{tabular}           & \begin{tabular}[c]{@{}c@{}} 77.18 $\pm$ 0.16\\ (0.88 $\times$)\end{tabular}           & \begin{tabular}[c]{@{}c@{}} \textit{77.14 $\pm$ 0.21}\\ \textit{(1.10 $\times$)}\end{tabular}           & \begin{tabular}[c]{@{}c@{}} 77.26 $\pm$ 0.18\\ (0.84 $\times$)\end{tabular}\\ \hline
\multirow{4}{*}[-1.5ex]{Tiny-ImageNet} & ResNet     & \begin{tabular}[c]{@{}c@{}}61.29 $\pm$ 0.41\\ (1 $\times$)\end{tabular}        & \begin{tabular}[c]{@{}c@{}}60.68 $\pm$ 0.30\\ (0.70 $\times$)\end{tabular}           & \begin{tabular}[c]{@{}c@{}}\textit{61.32 $\pm$ 0.30}\\ \textit{(1.02 $\times$)}\end{tabular}           & \begin{tabular}[c]{@{}c@{}}\textit{61.41 $\pm$ 0.27}\\ \textit{(1.16 $\times$)}\end{tabular}           & \begin{tabular}[c]{@{}c@{}}59.36 $\pm$ 0.48\\ (0.78 $\times$)\end{tabular}           & \begin{tabular}[c]{@{}c@{}}\textbf{\textit{61.56 $\pm$ 0.31}}\\ \textit{(1.10 $\times$)}\end{tabular}           & \begin{tabular}[c]{@{}c@{}}60.59 $\pm$ 0.19\\ (0.71 $\times$)\end{tabular}           \\
     & ConvNeXt   & \begin{tabular}[c]{@{}c@{}} 61.40 $\pm$ 0.59\\ (1 $\times$)\end{tabular}        & \begin{tabular}[c]{@{}c@{}} 62.37 $\pm$ 0.37\\ (0.84 $\times$)\end{tabular}           & \begin{tabular}[c]{@{}c@{}} \textit{61.55 $\pm$ 0.68}\\ \textit{(1.11 $\times$)}\end{tabular}           & \begin{tabular}[c]{@{}c@{}}\textit{61.42 $\pm$ 0.62}\\ \textit{(1.18 $\times$)}\end{tabular}           & \begin{tabular}[c]{@{}c@{}}62.13 $\pm$ 0.57\\ (0.88 $\times$)\end{tabular}           & \begin{tabular}[c]{@{}c@{}}\textit{61.80 $\pm$ 0.56}\\ \textit{(1.14 $\times$)}\end{tabular}           & \begin{tabular}[c]{@{}c@{}}\textbf{62.56 $\pm$ 0.53}\\ (0.85 $\times$)\end{tabular}           \\
     & Swin & \begin{tabular}[c]{@{}c@{}} 58.12 $\pm$ 0.13\\ (1 $\times$)\end{tabular}        & \begin{tabular}[c]{@{}c@{}}58.41 $\pm$ 0.06\\ (0.90 $\times$)\end{tabular}           & \begin{tabular}[c]{@{}c@{}} N/A \\ N/A\end{tabular}           & \begin{tabular}[c]{@{}c@{}}\textit{59.12 $\pm$ 0.23} \\ \textit{(1.16 $\times$)}\end{tabular}           & \begin{tabular}[c]{@{}c@{}}59.61 $\pm$ 0.08 \\ (0.93 $\times$)\end{tabular}           & \begin{tabular}[c]{@{}c@{}}\textit{59.64 $\pm$ 0.26}\\ \textit{(1.14 $\times$)}\end{tabular}           & \begin{tabular}[c]{@{}c@{}}
    \textbf{59.89 $\pm$  0.13}\\ (0.85 $\times$)\end{tabular}           \\ \toprule
\end{tabular}
}
    \caption{Top-1 accuracy comparison on different datasets and backbones. The value in parentheses is the speedup ratio compared to its normalizer baseline.
\textbf{Bold} denotes the highest accuracy, \textit{italic} denotes exceeding the normalizer baseline in both accuracy and speed.}
		\label{table:class}
\end{table*}

\section{Experiments}
\label{section: experiment}
Experiments in this paper are run on an Ubuntu 16.04 LTS server with  8$\times$NVIDIA Tesla P100 (16GB) GPUs. We implement all deep learning models based on  PyTorch 1.7.1 \cite{pytorch} with Cuda 10.2. We provide the core python implementation of NoMorelization in Appendix \ref{app:code}.
	\subsection{Assertions Tests}
	\label{sec:assertion}
\begin{figure}
		\centering
		\includegraphics[width=0.8\linewidth]{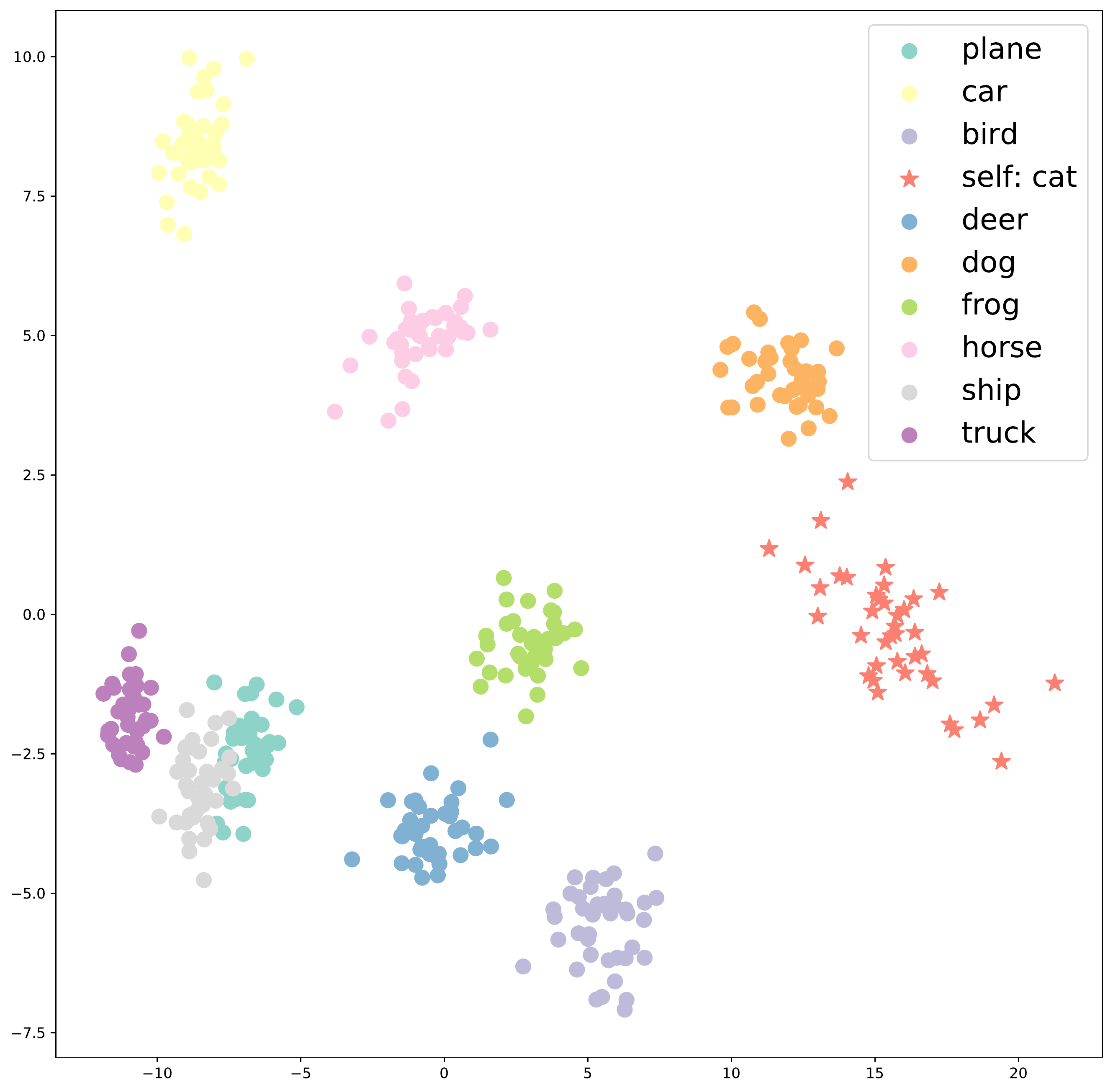}
		\caption{PCA results of the $\hat{\boldsymbol{x}_{i}}$ of the invariant cat sample within a batch of different distributions. The special case of inter-class noise and intra-class noise are shown vividly. }
		\label{fig:NoiseModeling}
	\end{figure}

    We select a cat picture as an invariant sample $x_i$ in CIFAR-10 \cite{cifar} dataset, and use Hotelling's $T^2$ hypothesis test \cite{hotelling} to validate our assertions. It is worth mentioning that we use a trained ResNet in the following assertions testing, so the \textbf{differences between different embedding results for the invariant sample are only due to the sampling noise in the batch}. See Appendix \ref{app:test} for more details and results.
    \paragraph{Assertion test 1}
    We divide the images of each category in the dataset into 40 parts and choose a category arbitrarily. The invariant samples are concatenated with 50 parts of pictures to form 40 batches.  By feeding the batches into a ResNet-56 model in training mode, we can get 40 random samples of 64-dim embedding vector of the invariant cat $\hat{\boldsymbol{x}_{i}}(1),\cdots,\hat{\boldsymbol{x}_{i}}(40)$. After that, we subtract the embeddings of different sampling indexes to get $C^2_{40} = 780$ sets of sampled intra-class noise. We perform a one-sample test between sampled intra-class noise and a zero matrix. 
    \paragraph{Assertion test 2}
    We set the batch size to 128 and randomly sample from the same class 1000 times, concatenate them with the invariant sample and then get 1000 different $\hat{\boldsymbol{x}_{i}}$ samples for ten classes. Moreover, we test them with a zero matrix respectively for hypothesis testing.
    \paragraph{Assertion test 3}
    We perform Principal Component Analysis (PCA) on $\hat{\boldsymbol{x}_{i}}$ for each class used in Assertion 1. As shown in Fig. \ref{fig:NoiseModeling}, the embedding shift of the invariant sample will be dominated by the inter-class noise especially when the samples are all of the same distribution (class).
	There is also slight intra-class noise within each class.

\subsection{Image Classification}
\paragraph{Datasets and baselines.}
We perform image classification experiments on four datasets, including three tiny image datasets (\textit{i.e., }CIFAR-10, CIFAR-100 \cite{cifar} and Tiny-ImageNet \cite{le2015tiny}) and a standard ImageNet \cite{imagenet} dataset.  To evaluate the generality of NoMorelization, we choose three types of backbone: ResNet-56 as a classical CNN with BN, ConvNeXt \cite{convnext} as a state-of-art CNN architecture with LN, and Swin-Transformer \cite{swin} as recently mainstream attention-based model. As far as we know, the current attention-based models all use LN. Please refer to Appendix \ref{app:class} for detailed model design and hyper-parameter settings.
\paragraph{Experimental Results.}  We train all models with different random seeds five times and compute the mean and standard deviation of top-1 accuracy. In order to compare the efficiency of different methods, we also record the average time-consuming of five training sessions. The accuracy and speed are recorded in Tab. \ref{table:class}. NoMorelization is the only method that can exceed the normalizer baseline in both accuracy and speed across all backbones and datasets. Furthermore, the highest accuracy can be obtained by NoMorelization in most cases with the additional regularization of NFNets, if training speed is not a concern. Finally, it is worth mentioning that Fixup initialization \cite{fixup} is designed for ResNet. Although Fixup initialization can achieve good results on CNN-based backbones, applying it to Transformer-based backbones will make the model difficult to converge.

\begin{figure}[!h]
		\centering
		\includegraphics[width=0.8\linewidth]{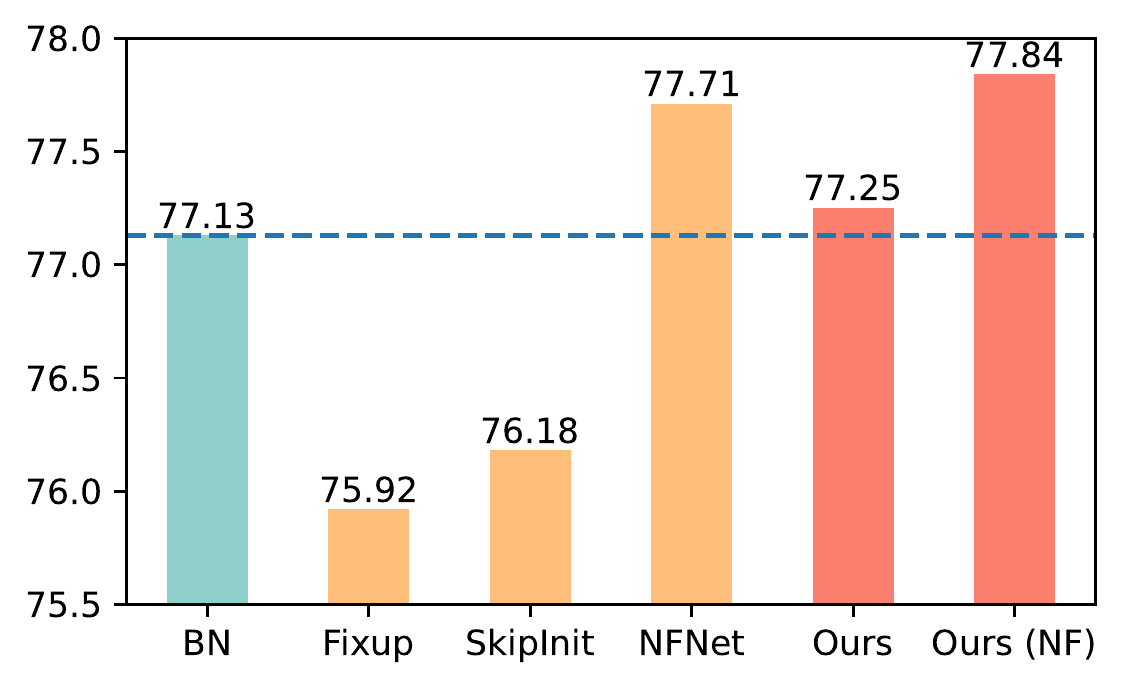}
		\caption{Top-1 Acc on ImageNet dataset with ResNet-50. Ours (NF) means we perform additional gradient clipping and weight standardization like NFNets. The green, yellow, and red bars represent the normalizer baseline, existing NF methods (\textit{i.e.,} Fixup, SkipInit \cite{skipinit}, and NFNet), and NoMorelization, respectively. }
		\label{fig:imagenet}
	\end{figure}

\paragraph{Evaluating on ImageNet.}
We also implement a standard ResNet-50 to evaluate the performance of different methods on a large-scale dataset. As shown in Fig. \ref{fig:imagenet}, only NFNet and our NoMorelizaion can outperform the BN baseline. The additional regularization of NFNet, as we mentioned in Sec. \ref{sec:why}, is not a replacement for BN but a complement. That means using NFNet-like gradient clipping and weight standardization along with our NoMorelization can improve performance and get the best results.

\begin{table}[!h]
\centering
\small
\begin{tabular}{ccccc}
\toprule
               & \multicolumn{2}{c}{summer2winter} & \multicolumn{2}{c}{horse2zebra}  \\ \cline{2-5} 
               & FID $\downarrow$  & IS $\uparrow$ & FID $\downarrow$ & IS $\uparrow$ \\ \hline
IN             & 83.72             & 2.78 & 64.52            & 1.42          \\
NoMorelization & \textbf{81.83}    & \textbf{2.83}          & \textbf{63.00}   & \textbf{1.50} \\ \toprule
\end{tabular}
    \caption{Quantitative comparison of generative tasks.}
		\label{table:generation}
\end{table}

\begin{figure*}[!h]
    \centering  
    \subfigure[Input horse]{
        \label{Fig.horse.1}        \includegraphics[width=0.153\textwidth]{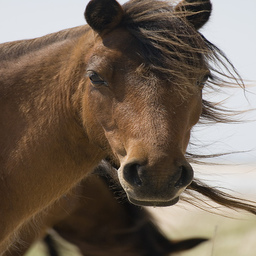}
        }
    \subfigure[IN]{
        \label{Fig.horse.in}
        \includegraphics[width=0.153\textwidth]{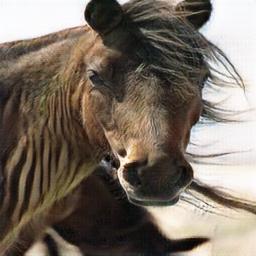}
        }
    \subfigure[NoMorelization]{
        \label{Fig.horse.nm}
        \includegraphics[width=0.153\textwidth]{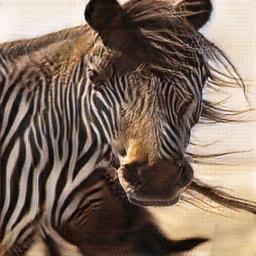}
        }
        \subfigure[Input summer]{
        \label{Fig.summer.1}
        \includegraphics[width=0.153\textwidth]{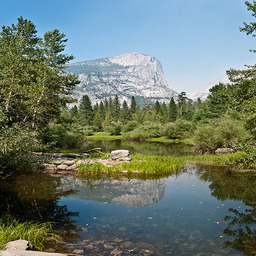}
        }
    \subfigure[IN]{
        \label{Fig.sumer.in}
        \includegraphics[width=0.153\textwidth]{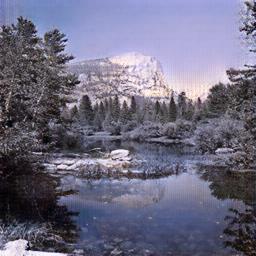}
        }
    \subfigure[NoMorelization]{
        \label{Fig.summer.nm}
        \includegraphics[width=0.153\textwidth]{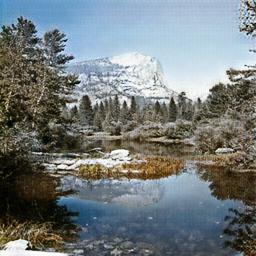}
        }    
    \caption{Qualitative comparison of generative tasks.}
    \label{Fig.real}
\end{figure*}
\subsection{Image-to-Image Translation}
To verify the effect of NoMorelization on generative tasks, we implement CycleGAN \cite{cycle} based on MMGeneration \cite{2021mmgeneration}. CycleGAN is a GAN model applied to image-to-image translation tasks. CycleGAN uses Instance Normalization by default. First, we train a standard CycleGAN model on the Summer-to-Winter and Horse-to-Zebra datasets\footnote{\url{https://people.eecs.berkeley.edu/~taesung_park/CycleGAN/datasets/}}. Then we replace IN in CycleGAN with NoMorelization. Finally, we perform quantitative and qualitative comparisons of CycleGAN using IN and NoMorelization. We train CycleGANs for 250k iterations. 
For quantitative comparison, we generate 128 images after every 10k iterations of training and calculate their Frechet Inception Distance (FID) \cite{FID} and Inception Score (IS) \cite{IScore}. Lower FID and higher IS indicate better results.  We report the lowest FID obtained by the models and the IS at this time, as shown in Tab. \ref{table:generation}. NoMorelization outperforms the IN baseline in quantitative scores under the same hyperparameter settings. For qualitative comparison, we run the CycleGAN checkpoint in Tab. \ref{table:generation} on the test set. Results are shown in Fig. \ref{Fig.real}. On two tasks, the generative models with NoMorelization perform well intuitively compared to the IN baselines. More NoMorelization-based generation results and training settings are in Appendix \ref{app:generate}.

	\begin{figure}[!h]
		\centering
        \subfigure[Sensitivity analysis on ResNet]{
            \label{Fig.ablation.1}
            \includegraphics[width=0.7\linewidth]{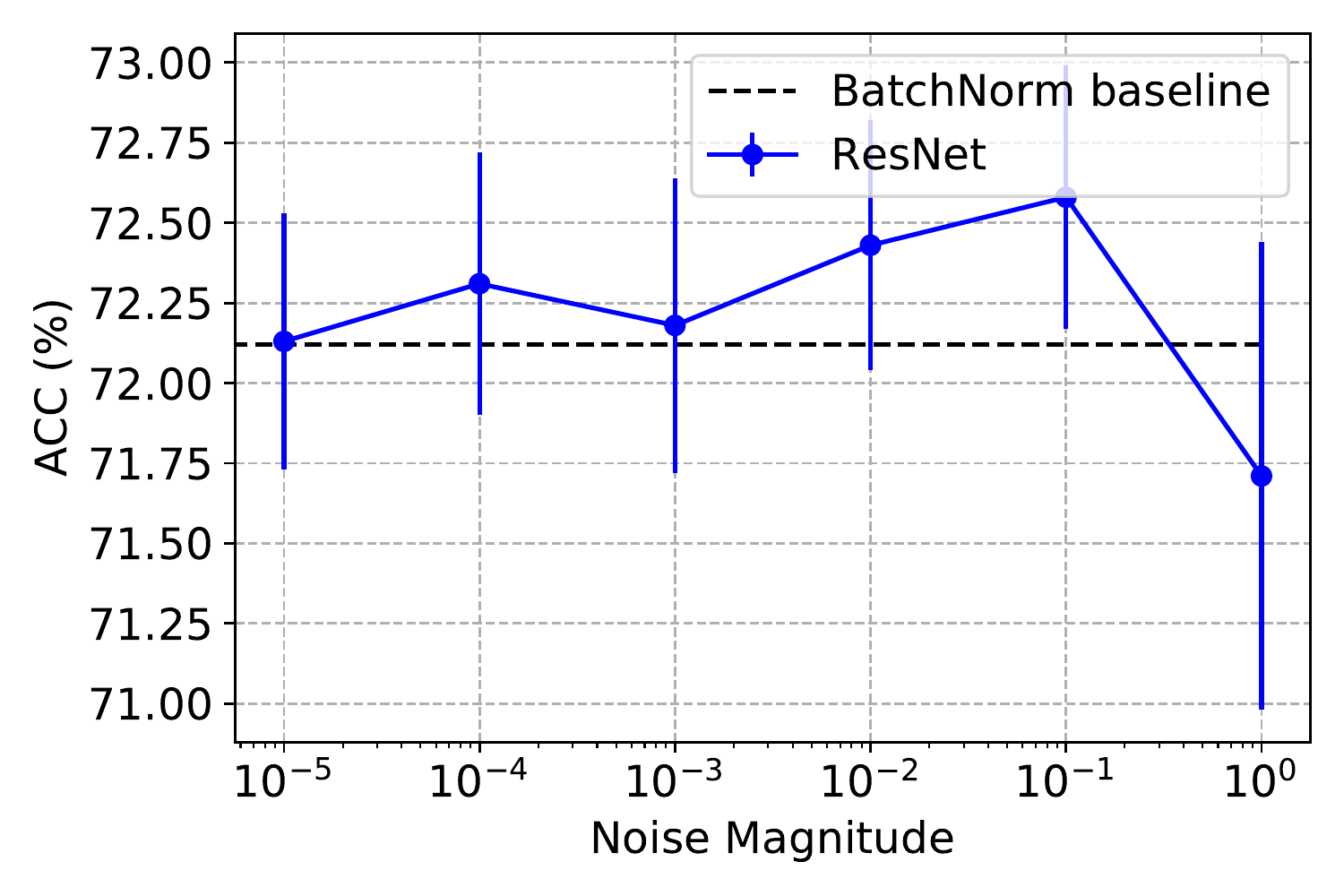}
            }
        \subfigure[Sensitivity analysis on Swin-Transformer]{
            \label{Fig.ablation.2}
            \includegraphics[width=0.7\linewidth]{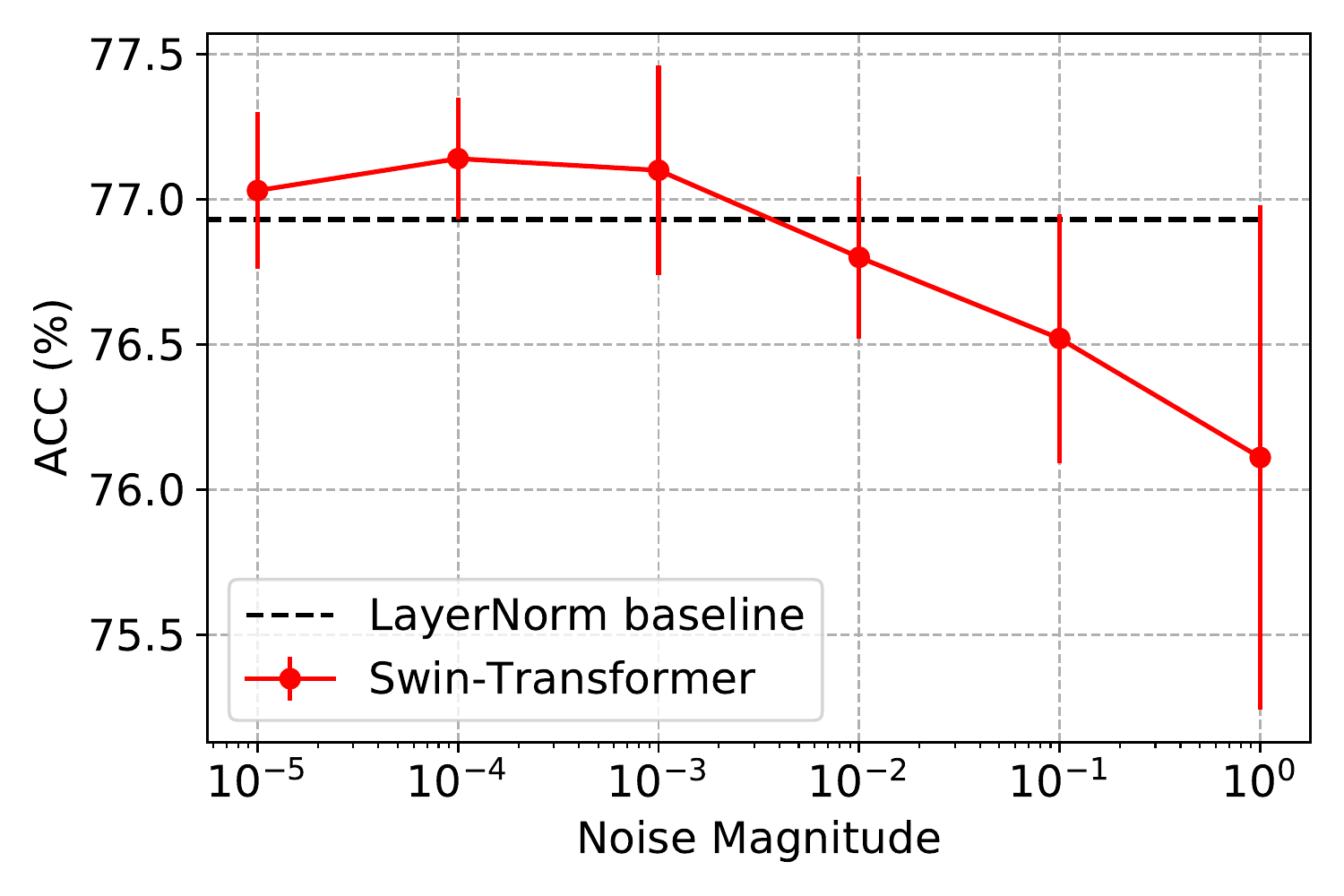}
            }
		\caption{Sensitivity analysis on noise amplitude. We choose ResNet and Swin-Transformer on the CIFAR-100 dataset as examples. The black line represents the normalizer baseline.} 
		\label{fig:ablation}
	\end{figure}
\subsection{In-depth Analysis}
\label{sec:ablation}
To determine the appropriate noise for different model architectures, we performed a log-scale grid search for magnitude among $[1, 1e-1, \cdots, 1e-5]$ on different backbones. As a result, we find that different backbones have different sensitivity of noise injected by NoMorelization.
BN-based backbones can tolerate more extensive noise (\textit{i.e.,} 1e-1), while LN-based backbones prefer smaller noise amplitudes (\textit{i.e.,} 1e-4).
We take the BN-based ResNet and LN-based Swin-Transformer as examples. As shown in Fig. \ref{Fig.ablation.1}, ResNet backbone with NoMorelization easily exceeds the BN baseline. Such advantages are more pronounced when the noise amplitude gradually increases from 1e-5 to 1e-1. However, if the magnitude of injected noise is too large (\textit{i.e.,} 1), the accuracy of NoMorelization will drop rapidly and make the training results unstable. For the Swin-Transformer shown in Fig. \ref{Fig.ablation.2}, although NoMorelization can achieve higher accuracy than its baseline, the accuracy drop at early stage. The noise starts to have negative effects on the model when its magnitude reaches 1e-2.  

The results of the sensitivity analysis also explain why we prefer BN as a normalizer for CNN-based architectures and LN for Transformer-based architectures.
As the only common normalizer that introduces sampling noise, BN brings a greater noise regularization effect than NoMorelization. For noise-sensitive architectures (such as Transformer), the regularization of BN is too strong. Therefore, normalizers that do not introduce noise regularization (\textit{e.g.,} LN and IN) are chosen when implementing these models.

\section{Conclusion}
In this paper, we propose a simple and effective alternative to normalization, namely ``NoMorelization'', by explaining BN's effects from a sample's perspective. 
NoMorelization is composed of two trainable scalars and a zero-centered noise injector.
The experimental results validate our assumptions about the injected noise and show that NoMorelization is a general component of deep learning and is suitable for different model paradigms to tackle different tasks. Furthermore, compared with existing mainstream normalizers and state-of-the-art normalizer-free methods, NoMorelization shows the best speed-accuracy trade-off.

\bibliography{aaai23}

\clearpage

\appendix

\onecolumn
\section{NoMorelization Implementation}
\label{app:code}
\subsection{Environment}
\paragraph{Hardware.} Experiments in this paper are run on an Ubuntu 16.04 LTS server with 8$\times$NVIDIA Tesla P100 (16GB) GPUs, Intel(R) Xeon(R) Gold 5115 CPU @2.40GHz, and 252 GB memory. 
\paragraph{Software.} We implement all deep learning models based on Python 3.8.10, PyTorch 1.7.1 \cite{pytorch} with Cuda 10.2, torchvision 0.8.2, mmcv 1.6.0,  mmgeneration 0.7.1, and their dependencies.

\subsection{Image Classification}
\label{app:block code}
For a fair comparison, we implement NoMorelization's experiment based on existing open source code. The original code can be found in the comments in the code block below.

\begin{python}[!h]
# NoMorelization ResNet Block
# Modified from https://github.com/hongyi-zhang/Fixup
class BasicBlock(nn.Module):
    def __init__(self, inplanes, planes, stride=1, downsample=None):
        super(BasicBlock, self).__init__()
        # Both self.conv1 and self.downsample layers downsample the input when stride != 1
        self.conv1 = conv3x3(inplanes, planes, stride)
        self.relu = nn.ReLU(inplace=True)
        self.conv2 = conv3x3(planes, planes)
        self.downsample = downsample
        self.alpha = nn.Parameter(torch.zeros(1))
        self.beta = nn.Parameter(torch.zeros(1))
        
    def forward(self, x):
        identity = x
        out = self.conv1(x) 
        out = self.relu(out)
        out = self.conv2(out)
        out = out * self.alpha + self.beta
        if self.training:    
            out += torch.randn_like(out, device=out.device) * 0.1        
        if self.downsample is not None:
            identity = self.downsample(x)
            identity = torch.cat((identity, torch.zeros_like(identity)), 1)
        out += identity
        out = self.relu(out)
        return out

\end{python}
\begin{python}[!h]
# NoMorelization ConvNeXt Block
# Modified from https://github.com/facebookresearch/ConvNeXt
class Block(nn.Module):
    def __init__(self, dim, drop_path=0., layer_scale_init_value=1e-6):
        super().__init__()
        self.dwconv = nn.Conv2d(dim, dim, kernel_size=7, padding=3, groups=dim)
        self.pwconv1 = nn.Conv2d(dim, 4 * dim, kernel_size=1)
        self.act = nn.GELU()
        self.pwconv2 = nn.Conv2d(4 * dim, dim, kernel_size=1)
        self.gamma = nn.Parameter(layer_scale_init_value * torch.ones((dim)), 
                                    requires_grad=True) if layer_scale_init_value > 0 else None
        self.alpha = nn.Parameter(torch.zeros(1))
        self.beta = nn.Parameter(torch.zeros(1))
        self.drop_path = DropPath(drop_path) if drop_path > 0. else nn.Identity()

    def forward(self, x):
        identity = x
        x = self.dwconv(x)
        x = self.pwconv1(x)
        x = self.act(x)
        x = self.pwconv2(x)
        if self.gamma is not None:
            x = self.gamma * x 
        x = x * self.alpha + self.beta
        if self.training:
            x += torch.randn_like(x, device=x.device) * 1e-4            
        x = identity + self.drop_path(x)
        return x
\end{python}
\begin{python}[!h]
# NoMorelization Swin-Transformer Block
# Modified from https://github.com/aanna0701/SPT_LSA_ViT
class SwinTransformerBlock(nn.Module):
    def __init__(self, dim, input_resolution, num_heads, window_size=7, shift_size=0,
                 mlp_ratio=4., qkv_bias=True, qk_scale=None, drop=0., attn_drop=0., drop_path=0.,
                 act_layer=nn.GELU, is_LSA=False):
        super().__init__()
        self.dim = dim
        self.input_resolution = input_resolution
        self.num_heads = num_heads
        self.window_size = window_size
        self.shift_size = shift_size
        self.mlp_ratio = mlp_ratio
        if min(self.input_resolution) <= self.window_size:
            self.shift_size = 0
            self.window_size = min(self.input_resolution)
        assert 0 <= self.shift_size < self.window_size, "shift_size must in 0-window_size"

        self.attn = WindowAttention(
            dim, window_size=to_2tuple(self.window_size), num_heads=num_heads,
            qkv_bias=qkv_bias, qk_scale=qk_scale, attn_drop=attn_drop, proj_drop=drop, is_LSA=is_LSA)

        self.drop_path = DropPath(drop_path) if drop_path > 0. else nn.Identity()
        mlp_hidden_dim = int(dim * mlp_ratio)
        self.mlp = Mlp(in_features=dim, hidden_features=mlp_hidden_dim, act_layer=act_layer, drop=drop)
        self.alpha = nn.Parameter(torch.zeros(1))
        self.beta = nn.Parameter(torch.zeros(1))
        if self.shift_size > 0:
            # calculate attention mask for SW-MSA
            H, W = self.input_resolution
            img_mask = torch.zeros((1, H, W, 1))  # 1 H W 1
            h_slices = (slice(0, -self.window_size),
                        slice(-self.window_size, -self.shift_size),
                        slice(-self.shift_size, None))
            w_slices = (slice(0, -self.window_size),
                        slice(-self.window_size, -self.shift_size),
                        slice(-self.shift_size, None))
            cnt = 0
            for h in h_slices:
                for w in w_slices:
                    img_mask[:, h, w, :] = cnt
                    cnt += 1

            mask_windows = window_partition(img_mask, self.window_size)  # N_w^2, window_size, window_size, 1
            mask_windows = mask_windows.view(-1, self.window_size * self.window_size)   # N_w^2, window_size, window_size
            attn_mask = mask_windows.unsqueeze(1) - mask_windows.unsqueeze(2)   # (N_w^2, 1, window_size, window_size) - (N_w^2, window_size, 1, window_size)
            attn_mask = attn_mask.masked_fill(attn_mask != 0, float(-100.0)).masked_fill(attn_mask == 0, float(0.0))
        else:
            attn_mask = None

        self.register_buffer("attn_mask", attn_mask)    # No parameter

    def forward(self, x):
        B, L, C = x.shape
        H = int(math.sqrt(L)) 
        shortcut = x
        x = x.view(B, H, H, C)
        # cyclic shift
        if self.shift_size > 0:
            shifted_x = torch.roll(x, shifts=(-self.shift_size, -self.shift_size), dims=(1, 2))
        else:
            shifted_x = x
        # partition windows
        x_windows = window_partition(shifted_x, self.window_size)
        x_windows = x_windows.view(-1, self.window_size * self.window_size, C)
        attn_windows = self.attn(x_windows, mask=self.attn_mask)
        attn_windows = attn_windows.view(-1, self.window_size, self.window_size, C)
        shifted_x = window_reverse(attn_windows, self.window_size, H, H)
        # reverse cyclic shift
        if self.shift_size > 0:
            x = torch.roll(shifted_x, shifts=(self.shift_size, self.shift_size), dims=(1, 2))
        else:
            x = shifted_x
        x = x.view(B, L, C)
        x = x * self.alpha + self.beta
        if self.training:
            x += torch.randn_like(x, device=x.device) * 1e-4
        x = shortcut + self.drop_path(x)
        x = x + self.drop_path(self.mlp(x))
        return x
\end{python}
\subsection{Image-to-Image Translation}
\begin{python}[!h]
# NoMorelization CycleGAN Block
# Modified from https://github.com/open-mmlab/mmgeneration/blob/master/mmgen/models/architectures/cyclegan/modules.py
class ResidualBlockWithDropout(nn.Module):
    def __init__(self,
                 channels,
                 padding_mode,
                 norm_cfg=dict(type='IN'),
                 use_dropout=True):
        super().__init__()
        use_bias = True
        block = [
            ConvModule(
                in_channels=channels,
                out_channels=channels,
                kernel_size=3,
                padding=1,
                bias=use_bias,
                norm_cfg=None,
                padding_mode=padding_mode)
        ]
        print(block[0].norm_cfg)
        self.alpha = nn.Parameter(torch.zeros(1))
        self.beta = nn.Parameter(torch.zeros(1))
        if use_dropout:
            block += [nn.Dropout(0.5)]

        block += [
            ConvModule(
                in_channels=channels,
                out_channels=channels,
                kernel_size=3,
                padding=1,
                bias=use_bias,
                norm_cfg=None,
                act_cfg=None,
                padding_mode=padding_mode)
        ]

        self.block = nn.Sequential(*block)

    def forward(self, x):
        if self.training:
            out = x + self.alpha * self.block(x) + self.beta + torch.randn_like(x, device=x.device) * 1e-4
        else:
            out = x + self.alpha * self.block(x) + self.beta
        return out
\end{python}

\section{Assertion Tests}
\label{app:test}
We use Hotelling's $T^2$ hypothesis test \cite{hotelling} to verify the first two assertions in Sec. \ref{sec:assertion}.
Hotelling's $T^2$ hypothesis test is a standard multivariate test that extends the single variable Student's t-test and is often used to compare the equality of means of two multivariate variables. 
\paragraph{Assertion Test 1}
Specificly, we perform a one-sample test between sampled intra-class noise and a zero matrix in Assertion Tests 1. We calculate and obtain 
the $P-value = 0.9511$, which means that at the 0.05 level, we \textit{cannot} reject the null hypothesis, which indicates that the mean of sampled intra-class noise is indeed a zero vector.
This proves that Assertion 1 holds.
\paragraph{Assertion Test 2}
Similarly to Assertion 1, in the verification of Assertion 2, we performed one-sample hypothesis tests between the noise and zero matrices in the case of fixing a certain image of the cat category, and all other samples belong to the same category. The results of the ten classes are shown in Tab. \ref{pvalue:a2}. There are eight classes of results with p-values less than 0.05, meaning the results are significant at the 0.05 level; in particular, there are three results less than 0.01 with highly significant results, so that in these eight groups of hypothesis tests, we \textit{can} reject the null hypothesis that the means of the corresponding noise are not a zero vector, as described in Eq. \eqref{a2:eq2}. On the other hand, the p-value for the cat class results is greater than 0.05, implying that we \textit{cannot} reject the null hypothesis, which indicates that the mean of sampled noise is indeed a zero vector, as described in Eq. \eqref{a2:eq1}. In contrast to our Assertion 2, the p-value of the result for the dog class is also greater than 0.05, and we conjecture that the strong similarity between the data of the cat and dog classes affects the result of assertion.

\begin{table}[!h]
\centering
\begin{tabular}{ccccccccccc}
\toprule
Class   & Cat(self)   & Plane & Car   & Bird  & Deer & Dog   & Frog  & Horse & Ship  & Truck \\ \hline
P-Value & 0.660 & 0     & 0.016 & 0.028 & 0    & 0.153 & 0.040 & 0.019 & 0.012 & 0     \\ \toprule
\end{tabular}
\caption{P-Value of Assertion 2.}
\label{pvalue:a2}
\end{table}

\section{Image Classification Training Details}
\label{app:class}
 \subsection{Model Design}
 \paragraph{CIFAR-10, CIFAR-100, and Tiny-ImageNet}
  For CNN-based architecture, we divide models into three stages. The input image is first expanded to the initial channel numbers by a 2D convolutional layer. Next, all convolutional layers are set with the corresponding padding to ensure that the spatial size of input and output remains unchanged. Finally, we downsample the input space by average pooling (kernel size is 2) and double the channel number after the second and third stages. The design of each block can be found in Appendix \ref{app:block code}. 
 After three stages, we downsample the feature tensor to a 1D vector using adaptive average pooling and a fully connected layer for classification. In short, from the input batch, first stage input, second stage input, third stage input, classification layer input, and the output batch are:
 $[N,3,H,W]\rightarrow [N,C,H,W]\rightarrow [N,C,H,W]\rightarrow [N,2C,H/2,W/2]\rightarrow [N,4C,H/4,W/4]\rightarrow [N,4C]\rightarrow [N,\text{Num class}]$ , where $N, C, H, W$ ,and Num class are batch size, base channel number, image height, image width, and the number of classes, respectively.
The value of $C$ is different for different datasets. $C=16$ for CIFAR-10, $C=32$ for CIFAR-100, and $C=64$ for Tiny-ImageNet.
The number of blocks in each stage of ResNet and ConvNeXt are [9, 9, 9] and [3, 3, 3], respectively. 
Our Swin-Transformer is a direct reference to the existing design\footnote{\url{https://github.com/aanna0701/SPT_LSA_ViT/blob/main/models/swin.py}}.
 \paragraph{ImageNet}
 Our ResNet-50 on the ImageNet dataset uses the same design in the PyTorch model zoo\footnote{\url{https://pytorch.org/vision/main/models/generated/torchvision.models.resnet50.html}}.
 \subsection{Hyper-parameter Setting}
  We train for 200 epochs for each architecture and set the label smoothing to 0.1 to stabilize training. The remaining hyper-parameters vary for different backbones, but we keep the same hyper-parameter settings (such as learning rate, data augmentation, and weight decay) for the same backbone.
  \paragraph{CIFAR-10, CIFAR-100, and Tiny-ImageNet}
 The batch size for all architecture is all set to 128. 
 We use the SGD optimizer with momentum 0.9 and set the weight decay to 1e-5. We set the basic learning rate to 5e-2 for CNN-based models and 5e-4 for Swin-Transformer. We perform warm-up to linearly increase the learning rate from 1e-4 to the basic learning rate for the first ten epochs. For CNN-based models, we divide the learning rate by ten at the 100th and 150th epochs. For Swin-Transformer, we perform a cosine learning rate schedule. The following code implements data augmentation. 
 
 \begin{python}
import torchvision.transforms as transforms
# DATA Augmentation for CIFAR 10 
transform_train = transforms.Compose([
    transforms.RandomResizedCrop(32),
    transforms.RandomHorizontalFlip(),
    transforms.ToTensor(),
    transforms.Normalize((0.4914, 0.4822, 0.4465), (0.2023, 0.1994, 0.2010)),
    ])
 \end{python}
The differences between the different datasets are as follows:
\begin{itemize}
    \item CIFAR-10 is normalized with (0.4914, 0.4822, 0.4465), (0.2470, 0.2435, 0.2616). The image size is 32$\times$32.
    \item CIFAR-100 is normalized with (0.5070, 0.4865, 0.4409), (0.2673, 0.2564, 0.2762). The image size is 32$\times$32.
    \item Tiny-ImageNet is normalized with  (0.4802, 0.4481, 0.3975), (0.2770, 0.2691, 0.2821). The image size is 64$\times$64.
\end{itemize}
 Especially for ConvNeXt, all datasets are resized to 64×64. Additional data augmentations (\textit{e.g.}, mixup and auto augmentation, \textit{e.t.c.}) are used in the training of Swin-Transformer to ensure a fair comparison with the baseline. Details of augmentation settings can be found in default settings in the open source code\footnote{\url{https://github.com/aanna0701/SPT_LSA_ViT/blob/main/main.py}}. 
 \paragraph{ImageNet}
  The batch size of ResNet-50 is 64. The learning rate is set to 0.05 with a cosine scheduler. We use the SGD optimizer with momentum 0.9 and set the weight decay to 1e-5. RandAug\footnote{\url{https://github.com/rwightman/pytorch-image-models/blob/master/timm/data/auto_augment.py}} is implemented as data augmentation.

\section{Image-to-Image Translation Details}
\label{app:generate}
\subsection{Generation Results}
\begin{figure*}[!h]
    \centering 
    \subfigure{
        \label{Fig.more.1}        \includegraphics[width=0.23\textwidth]{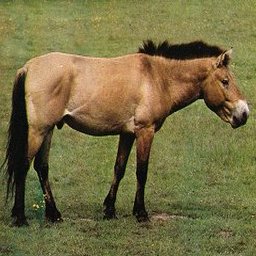}
        }
    \subfigure{
        \label{Fig.more.2}
        \includegraphics[width=0.23\textwidth]{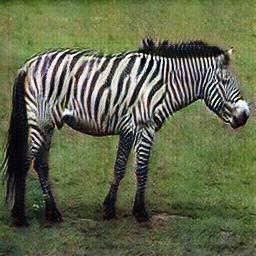}
        }
    \subfigure{
        \label{Fig.more.3}
        \includegraphics[width=0.23\textwidth]{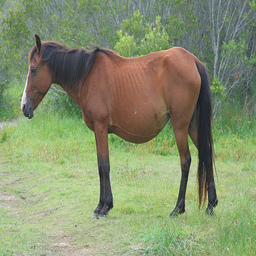}
        }
        \subfigure{
        \label{Fig.more.4}
        \includegraphics[width=0.23\textwidth]{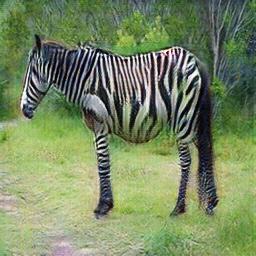}
        }
    \subfigure{
        \label{Fig.more.5}
        \includegraphics[width=0.23\textwidth]{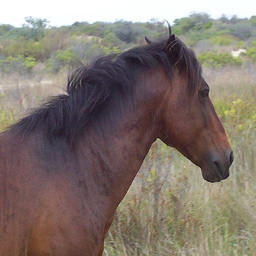}
        }
        \subfigure{
        \label{Fig.more.6}
        \includegraphics[width=0.23\textwidth]{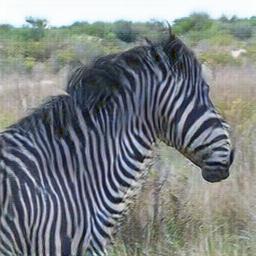}
        }     
    \subfigure{
        \label{Fig.more.7}
        \includegraphics[width=0.23\textwidth]{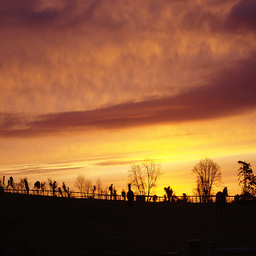}
        }
        \subfigure{
        \label{Fig.more.8}
        \includegraphics[width=0.23\textwidth]{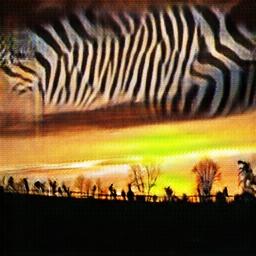}
        }                
    \caption{More Results of NoMorelization CycleGAN (Horse2Zebra).}
    \label{Fig.morereal}
\end{figure*}

\begin{figure*}[!h]
    \centering  
    \subfigure{
       \includegraphics[width=0.23\textwidth]{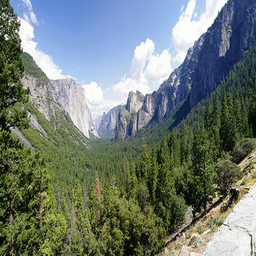}
        }
    \subfigure{
        \includegraphics[width=0.23\textwidth]{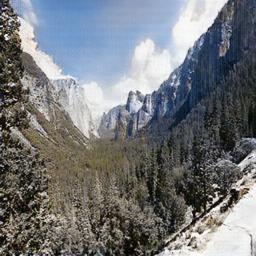}
        }
    \subfigure{
        \includegraphics[width=0.23\textwidth]{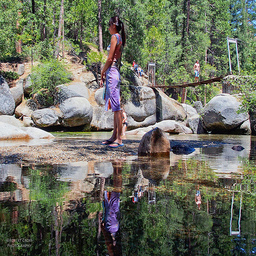}
        }
        \subfigure{
        \includegraphics[width=0.23\textwidth]{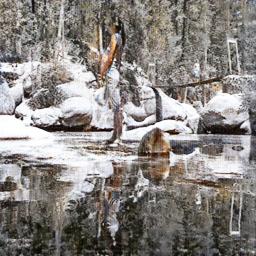}
        }
    \subfigure{
        \includegraphics[width=0.23\textwidth]{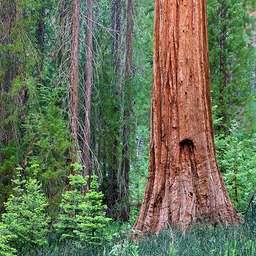}
        }
        \subfigure{
        \includegraphics[width=0.23\textwidth]{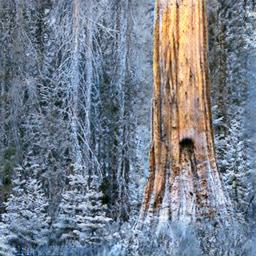}
        }     
    \subfigure{
        \includegraphics[width=0.23\textwidth]{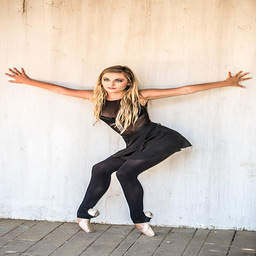}
        }
        \subfigure{
        \includegraphics[width=0.23\textwidth]{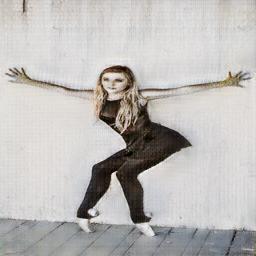}
        }                
    \caption{More Results of NoMorelization CycleGAN (Summer2Winter).}
    \label{Fig.morereal2}
\end{figure*}

We provide more generative results on NoMorelization CycleGAN in Fig. \ref{Fig.morereal} and Fig. \ref{Fig.morereal2}. Each of the two pictures is the original picture and the translated output of CycleGAN.
In the last example of each task, we also show the response of the model trained by NoMorelization when the input image does not have subjects in the training set.

\subsection{Hyper-parameter Setting}
We first scale the training images to 286$\times$286 by bicubic interpolation and then crop them to 256$\times$256. We randomly flip the training images horizontally with 50\% probability and finally normalize them by (0.5,0.5,0.5),(0.5,0.5,0.5). We use the Adam optimizer with a learning rate of 2e-5, a beta of (0.5, 0.999), and a linear learning rate schedule. We take Cycle Loss and Identity Loss with a weight of 0.5 as the criterion. The basic channel number of CycleGAN is 64. There are nine residual blocks in total, and the implementation of residual blocks can be found in Appendix \ref{app:code}.

\end{document}